\documentclass[sigconf]{acmart}
\usepackage{subcaption}
\usepackage{multicol}
\usepackage{graphicx}

\AtBeginDocument{%
\providecommand\BibTeX{{%
\normalfont B\kern-0.5em{\scshape i\kern-0.25em b}\kern-0.8em\TeX}}}

\setcopyright{acmcopyright}
\copyrightyear{2020}

\acmConference[NICE'20]{NICE '20: The 8th Annual Neuro-Inspired Computational Elements (NICE) Workshop}{March 17--20, 2020}{Heidelberg, Germany}
\acmBooktitle{NICE '20: The 8th Annual Neuro-Inspired Computational Elements (NICE) Workshop,
March 17--20, 2020, Heidelberg, Germany}

\begin{document}

\title{Critical Limits in a Bump Attractor Network of Spiking Neurons}

\author{Alberto Arturo Vergani}
\authornote{Both authors contributed equally to this research.}
\email{a.vergani@mdx.ac.uk}
\author{Christian Robert Huyck}
\authornotemark[1]
\email{c.huyck@mdx.ac.uk}
\affiliation{%
\institution{Middlesex University}
\streetaddress{The Burroughs, Hendon}
\city{London}
\state{United Kingdom}
\postcode{NW4 4BT}
}

%
%
%
%
%
%

\renewcommand{\shortauthors}{Vergani and Huyck}

\begin{abstract}

A bump attractor network is a model that implements a competitive
neuronal process emerging from a spike pattern related to an input
source. Since the bump network could behave in many ways, this paper explores some critical limits of the parameter space using various positive and negative weights and an increasing size of the input spike sources
The neuromorphic simulation of the bump-attractor network shows that it exhibits a stationary, a splitting and a divergent spike pattern, in relation to different sets of weights and input windows. The balance between the values of positive and negative weights is important in determining the splitting or diverging behaviour of the spike train pattern and in defining the minimal firing conditions.

\end{abstract}




\maketitle

\section{Introduction}
The bump attractor network is a biologically inspired network able to
describe several functions of the brain. Formally, it is a set of recurrently connected nodes (neurons) that have a stable pattern due to their time dynamics. The bump attractor is a particular case of an attractor network \cite{Eliasmith:2007} in which the behaviour is stationary instead of cyclic or chaotic. This is why it is also called a stationary bump. 

In computational neuroscience, there are also
several types of attractor neural networks that are linked with
specific brain functions, e.g., memorization, action planning,
recognition and classification. The use of the attractor network
framework permits the application of theories and methodologies of dynamical systems that can  investigate their characteristics (e.g.,  
stability vs. instability, convergence vs. divergence, stationary vs. non-stationary, etc.).

The brain processes multi-modal
information coming from different sources of inputs. Although there
are several modalities of stimuli (within the subject himself such as
thinking and recalling, reasoning, or from outside of the body as
the general perception of the sensory inputs), the brain selects and
analyses this huge amount of information, that is often ambiguous,
fragmented, and noisy. Therefore, the brain should make a decision and capture information that is relevant and use it to make adaptive actions.

In the brain, a decision is represented as a feature selection process in which a particular cell or group of cells fires, suppressing nearby cells. The winner take all model (WTA) \cite{maass2000computational} is a system able to select between many options, in which several neurons compete when an input is presented and then only one wins. The bump-like network behaviour is an example of the WTA neural functionality \cite{somers1995emergent,laing2001stationary}, since a group of correlated firing neurons, given a input, can be observed as the winners of the competition.  It can be implemented by a specific balance of
excitatory and inhibitory synapses (see for example surrounding
inhibition properties in a network of spiking neurons in Chen
\cite{chen2017mechanisms}, and as historical examples, the works on the patterns of a stable grid in Wilson \cite{wilson1973mathematical} and on the process of neuronal
selectivity in Edelman \cite{edelman1987neural}).

There is a lot of evidence that excitatory cells, that are
principal neurons, are associated with specialised inhibitory cells,
including interneurons or secondary cells, that are connected
with principal cells as well as other interneurons. Negative
feedback from the inhibitory cells modulates the proper dynamics of
the neuronal network, making a start-and-stop functioning in the
neuronal network in a specific group or sub-group of the
brain. Thinking about the opposite situation, if there were only excitatory
neurons, their positive spikes could lead to an excitation that
produces more excitation, leading to an avalanche effect that potentially
becomes simulated epilepsy or other pathologies related to too much
activity of neurons. From the other point of view, observing too
much inhibition could allow a brain network weakly firing and
not reacting sufficiently to external demands. Therefore, as a
generalization, the models derived from the cell assemblies hypothesis
could not be applied since they need activation of
transiently spiking groups of co-firing neurons.

Cell assembly is the term coined by a Donald Hebb in the 1949 to describe co-activated firing neurons during a mental process \cite{hebb1949}. The learning mechanism proposed by Hebb is the so-called hebbian learning rule, in which the strength of synapses depends on the spike persistence between presynaptic and postsynaptic cells. The hebbian learning is often defined with the slogan "Cells that fire together wire together" (see work by Koroutchev at al \cite{koroutchev2006improved} for an example of that learning studied with attractor neural network).

Other learning processes in bump-attractor networks have been studied recently by Seeholzer \cite{seeholzer2019stability}. They investigated the stability of the working memory in continuous attractor networks under the control of short-term plasticity. The short-term synaptic plasticity of recurrent synapses influences the continuous attractor systems since short-term facilitation stabilizes memory retention and short-term depression increases continuous attractor volatility. Using mutual information they evaluated the combined impact of the short-term facilitation and depression on the capacity of the network to retain stable working memory. The facilitation processes decrease both diffusion and directed drifts, while short-term depression tends to increase both.

This work does not investigate the learning processes in the network. A set of static weights were selected setting up the topology of the bump-attractor network with several possible configurations. Considering the role of positive and negative connections in structuring the behaviour of a spiking bump network, this work 
investigates what are the weights combinations that allow the
emergence of a stationary bump, and the combinations that, instead,
allow other kinds of pattern in the spike
trains. 

Taking into account the size of the signals that trigger 
the neural network, what are the critical limits that
induce different emergent behaviour of the bump attractor?
The critical points are: first, the minimal source of inputs able
to ignite the network; second, the minimal number of input sources
that determine the emergence of different patterns (or different
attractors).

In the next sections there will be the description of i) structure of the bump-network used, ii)the neuronal model used for neuromorphic hardware selected to execute the computational experiments, iii) presentation of the results and, finally, iv) their discussion with limits and future directions.

\section{The Bump-attractor network}\label{secBAN}
\begin{figure}[tbph!]
\centering
\includegraphics[width=0.7\linewidth]{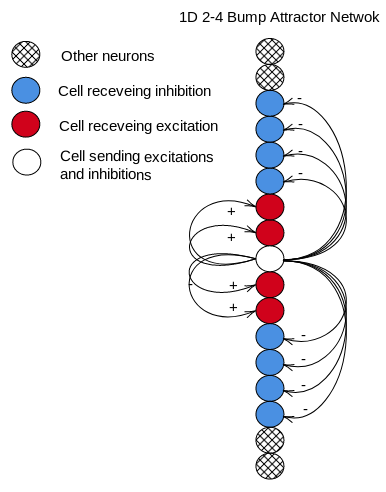}
\caption{Minimal topology of a 1D bump attractor network with 2 positive connections (excitatory synapses) and 4 negative connections (inhibitory synapses). Given a neuron, it sends excitation to the two nearest top and bottom neurons and inhibition to the four subsequent top and bottom neurons.}
\label{bumpNetwork}
\end{figure}

The structure of the bump-attractor is a 2-4 topology, where each neuron
has positive connections to the nearest two neurons on both sides ($ d<=2 $) and
negative connections to the next nearest 4 neurons on both sides ($
3<=d<=6 $) (see Figure \ref{bumpNetwork} for a minimal topology of the 1D 2-4 bump network). 

\begin{figure}[tbph!]
\centering
\includegraphics[width=0.8\linewidth]{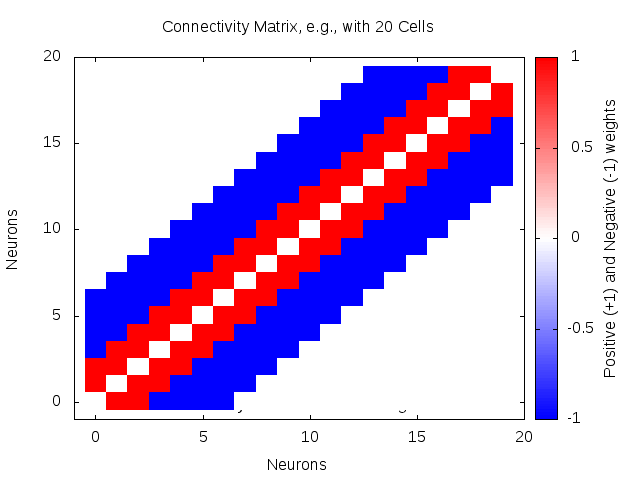} 
\caption{Connectivity matrix for a 1D bump attractor network of 20 cells. The weights are represented with  unitary values, e.g., positive weights equal to 1 and negative weights equal to -1.}.
\label{connectivityMatrix}
\end{figure}

Given N neurons, the connectivity matrix of a 1D 2-4 bump-attractor network is a squared matrix $A_{n,m} \in \mathbb{R}$ with $n=m=N$:  

\begin{equation}
A_{n,m} = 
\begin{pmatrix}
a_{1,1} & a_{1,2} & \cdots & a_{1,m} \\
a_{2,1} & a_{2,2} & \cdots & a_{2,m} \\
\vdots  & \vdots  & \ddots & \vdots  \\
a_{n,1} & a_{n,2} & \cdots & a_{n,m} 
\end{pmatrix}
\end{equation}{}

\noindent 
where $a_{n,m}$ is equal to 1 if it represents a positive weight (excitatory synapse) or -1 if it represent a negative weight (inhibitory synapse). Note that $a_{n,m}$ is equal to 0 when there are no connections: in case of the 2-4 topology of the bump-attractor network are self connections or connections beyond the 2-4 boundary. Figure \ref{connectivityMatrix} shows an example of a connectivity matrix for a bump-attractor network having 2-4 topology with 20 neurons.
%
%
%
%
%


This work explores the number of weights range from 0.05 to 0.10 (step equal to 0.01). The combinationS between positive and negative weights is $ 6 \times 6 = 36 $. The input windows of spike sources range from 30 to 70 (step equal to 1), obtaining in total $70-30=40$ possible firing inputs. Therefore, there are $36 \times 40 = 1440$ connectivity matrices. 

The next section describes the neuronal model selected for the bump-attractor network and the hardware architecture adopted to run the experiments.

\section{Simulations}
The computations has been made selecting first a neuronal model to use in the bump-attractor network (Subsection 3.1) and then running the simulation in a dedicated neuromorphic environment (Subsection 3.2).

\subsection{The neuronal model}
The model of the biological neuron used in the simulation is the leaky
integrate and fire model with a fixed threshold\footnote{The exponential integrate-and-fire neuron model is a particular case
of the AdEx model by removing the adaptation current ($ w $)
\citep{Gerstner:2009}.}. Synaptic conductance
is transmitted at a decaying-exponential rate from the pre to
post-synaptic neurons \citep{gerstner2014neuronal}.  The mathematical
description of the model follows the work of Fourcaud
\cite{fourcaud2003spike} (see also
\citep{richardson2003conductance,gewaltig2007nest}).

Equation \ref{dvmdt} describes the temporal changes of the
potential. The activation is the membrane potential $V_M $ and $C_M$
is the membrane capacity.  The four currents are the leak current, the
currents from excitatory and inhibitory synapses, and the input
current (from an external spike source).  The variable currents are
governed by equations \ref{isynex}, \ref{isynin} and \ref{il}.  In
equations \ref{isynex} and \ref{isynin} $E^{rev}_{Ex}$ and
$E^{rev}_{In}$ are the reversal potentials; excitation and inhibition
change slow as the voltage approaches these reversal potentials. In
equation \ref{il}, $V_{rest}$ is the resting potential of the neuron,
and $\tau_M$ is the leak constant.

\begin{equation}\label{dvmdt}
\frac{dV_M}{dt} =  \frac{( -I_{Leak} - I^{syn}_{Ex} - I^{syn}_{In} + I_{Ext} )}{C_M}
\end{equation}

\begin{equation}\label{isynex}
I^{syn}_{Exc} = G_{Ex} \times ( V_M - E^{rev}_{Ex} )
\end{equation}
\begin{equation}\label{isynin}
I^{syn}_{Inh} = G_{In} \times ( V_M - E^{rev}_{In} )
\end{equation}
\begin{equation}\label{il}
I_{Leak} = \frac { C_M \times (V_M - V_{rest})}{\tau_M}
\end{equation}

\begin{equation}\label{gex}
G_{Ex}(t) = k_{Ex} \times t \times e^{-\frac{t}{\tau^{syn}_{Ex}}}
\end{equation}
\begin{equation}\label{gin}
G_{In}(t) = k_{In} \times t \times e^{-\frac{t}{\tau^{syn}_{In}}}
\end{equation}

In the Equations \ref{gex} and \ref{gin}, the $ G_{Ex} $ and $ G_{In}
$ are the conductance in siemens to scale the post-synaptic potential
amplitudes used in equation \ref{isynex} and \ref{isynin}. $t$ is the
time step.

The constant $ k_{Ex} $ and $k_{In}$ are chosen so that $
G_{Ex}(\tau^{syn}_{Ex})=1 $ and $ G_{In}(\tau^{syn}_{In})=1 $. The $
\tau^{syn}_{Ex} $ and the $ \tau^{syn}_{In} $ are the decay rate of
excitatory and inhibitory synaptic current. 

When the voltage reaches the threshold, there is a spike and the voltage
is reset. No current is transferred during the refractory period.
In these simulations $ v_{thresh} = -48.0 $mV,
$\tau_{refract} = 2.0 $ ms.  The time step $t$ is 1ms.
$C_M = 1.0$nF, $ v_{reset}= -70.0$mV, $ v_{rest} = -65.0 $mV, $E^{rev}_{Ex}=0.0 $mV,
$E^{rev}_{In} = -70$mV, $\tau^{syn}_{Ex} = 5.0$ms, $\tau^{syn}_{In} = 5.0$ms
and $\tau_M = 20.0$ms.  These are all the default values.
The particular parameters
$ v_{thresh}$,$\tau_{refract}$, and  $t$, were selected as the authors have used
them for prior simulations.

\subsection{The computational setting}
The bump-attractor network of leaky integrate and fire neurons has
been implemented on SpiNNaker neuromorphic hardware
\cite{furber2014spinnaker}. The whole SpiNNaker system is a spiking
neural network architecture designed to deliver a massively parallel
million-core computer whose interconnecting architecture is inspired by
the connectivity characteristics of the mammalian brain. For the
purpose of this paper, we adopted the 4-chip board that has 72 ARM
processor cores, which will typically be deployed as 64 application
cores, 4 Monitor Processors and 4 spare cores. The simulations are
run with PyNN \citep{davison2007pynn} to specify the topology,
model, type of inputs, and recording of neuronal states.

Given the experimental conditions described in the above sections, the computational experiments is describable with the following pipeline:
\begin{itemize}
\item set-up the topology of a 1D bump-attractor network (2-4 connectivity) with 100 cells;   
\item simulation of the bump-attractor network dynamics given a
specified set of deterministic spike sources, ranging from 1 to 40
inputs, using a unitary step; in other words, the window of inputs is from
1 spike source (window 31-30=1) to 40 spike sources (window 70-30=40). 
\item the computational run-time  used is 300ms;
\item simulations have been run with a different combination of
positive and negative weights, varying from 0.05 to 0.10 (step equal
to 0.01);
\item the questions that are investigated are 1) if the network ignites
and 2) if it does, do the spike trains have either a stable
persistence, a splitting shape or a divergent pattern?;
\item in particular, the splitting spike behaviour has been evaluated
taking into account the number of streams, i.e., 2, 3 and 4, and the
possible combination with the divergence pattern.
\end{itemize}

The next section presents the results achieved.

\begin{figure*}
	\begin{multicols}{2}
		\includegraphics[width=\linewidth]{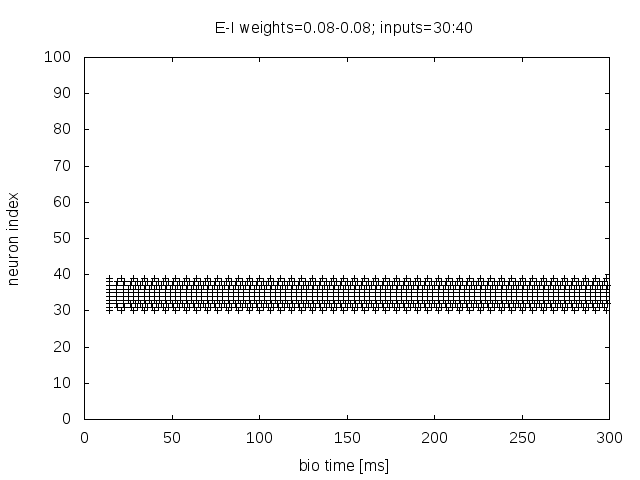} (a)
		
		\includegraphics[width=\linewidth]{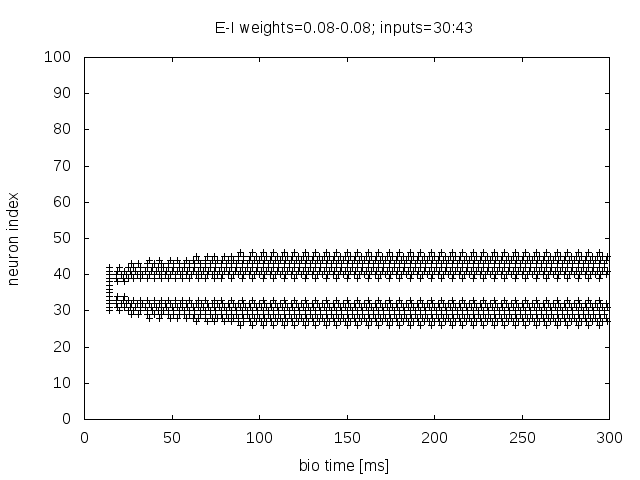} (b)
	\end{multicols}
	\begin{multicols}{2}
		
		\includegraphics[width=\linewidth]{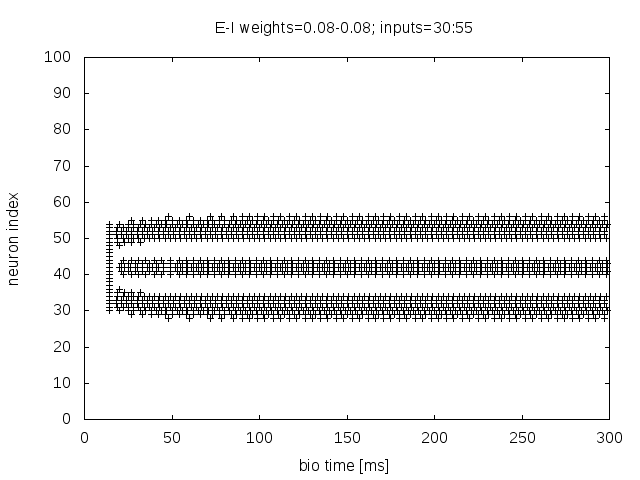} (c)
		\includegraphics[width=\linewidth]{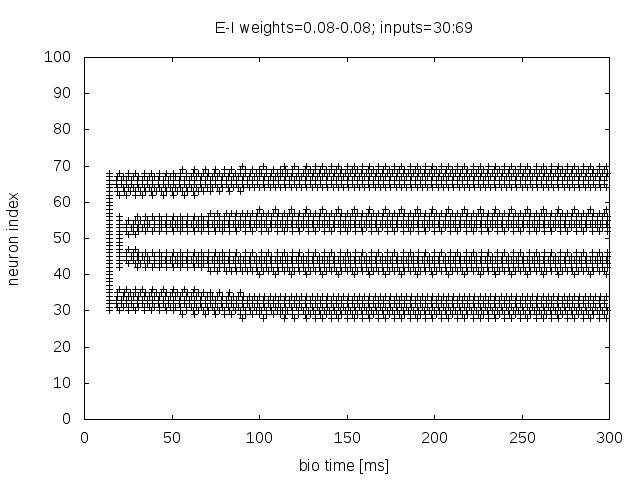} (d)
	\end{multicols}
	\begin{multicols}{2}
		
		\includegraphics[width=\linewidth]{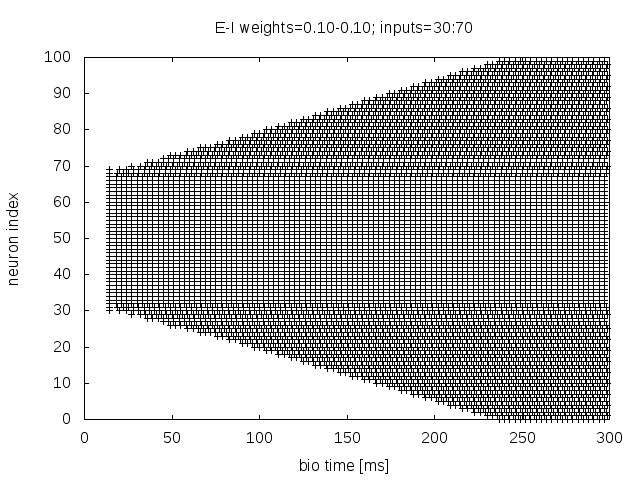} (e)
		
		\includegraphics[width=\linewidth]{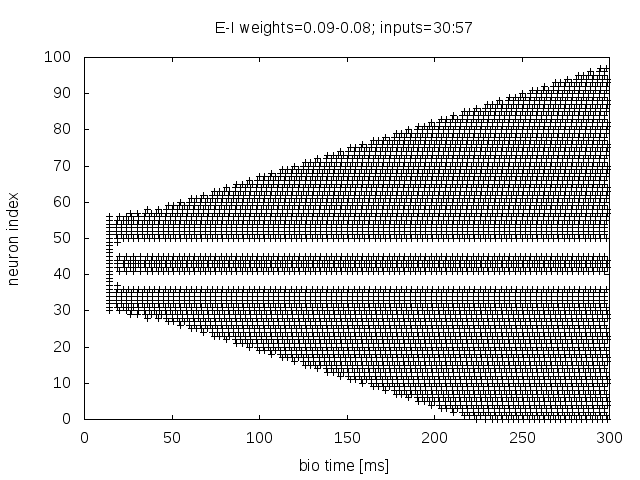} (f)
	\end{multicols}
	\caption{The figure shows the six different spike train
		patterns obtained with the simulations of a bump attractor
		network having different width of window inputs and
		positive and negative weight combinations. The (a) plot is 
		stationary persistence of spikes. The (b) plot is the
		splitting behaviour with two streams of spikes. The (c) is
		the splitting with three streams. The (d) plot is the
		splitting with four streams. The (e) is the diverging spike
		train pattern. The (f) is the splitting behaviour with
		divergence. See also Figure \ref{voltage} to see the relative voltage potential variations.}
	\label{spikes}
\end{figure*}

\begin{figure*}
	\begin{multicols}{2}
		\includegraphics[width=\linewidth]{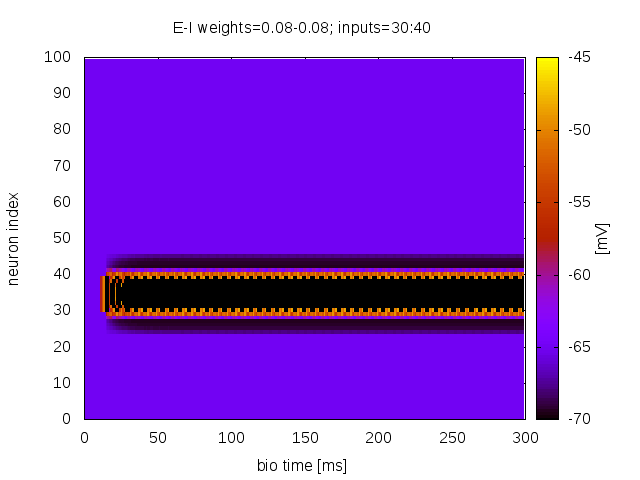} (a)
		
		\includegraphics[width=\linewidth]{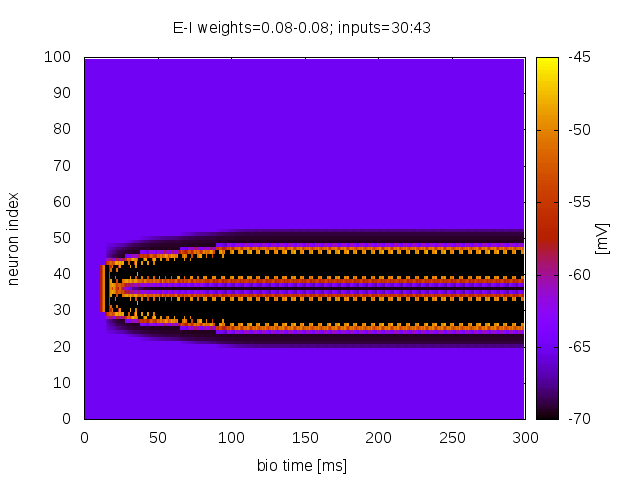} (b)
	\end{multicols}
	\begin{multicols}{2}
		
		\includegraphics[width=\linewidth]{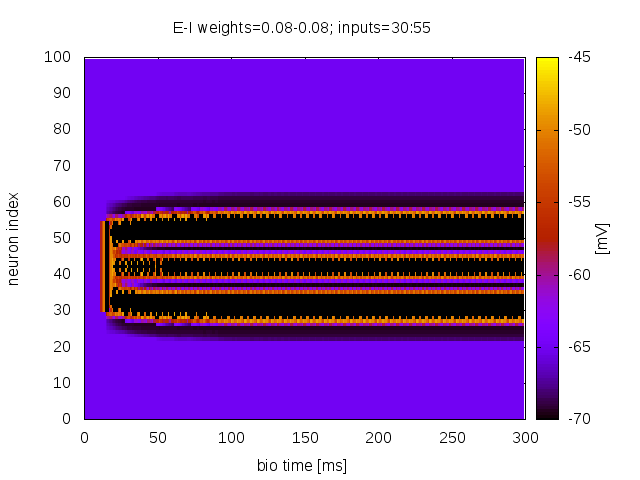} (c)
		\includegraphics[width=\linewidth]{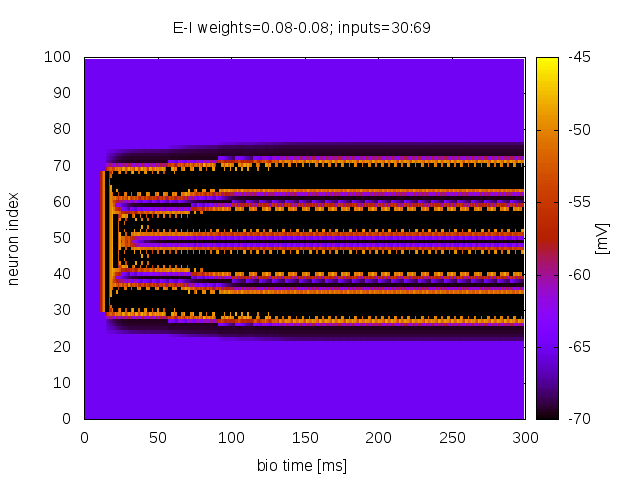} (d)
	\end{multicols}
	\begin{multicols}{2}
		
		\includegraphics[width=\linewidth]{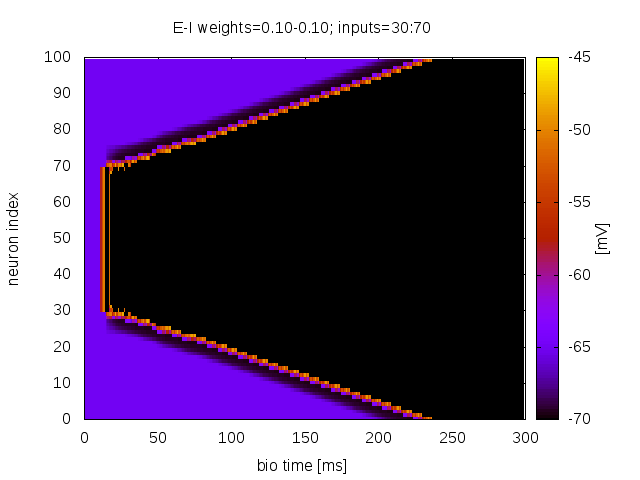} (e)
		
		\includegraphics[width=\linewidth]{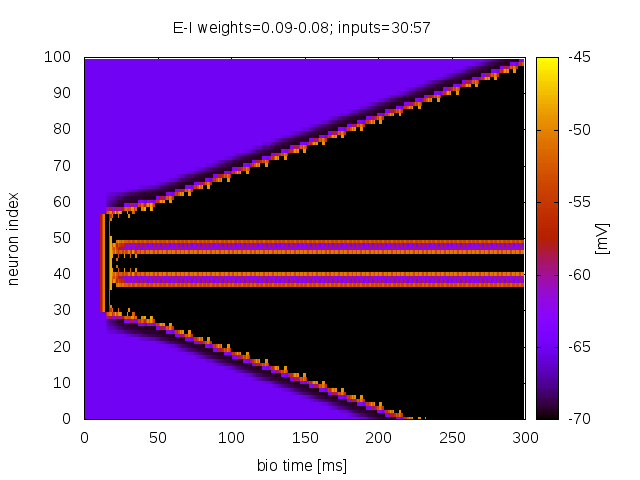} (f)
	\end{multicols}
	\caption{The figure shows the voltage potential variability associated to the six different spike train
		patterns obtained with the simulations of a bump attractor
		network having different size of window inputs and
		positive and negative weight combinations. The (a) plot is 
		stationary persistence of spikes. The (b) plot is the
		splitting behaviour with two streams of spikes. The (c) is
		the splitting with three streams. The (d) plot is the
		splitting with four streams. The (e) is the diverging spike
		train pattern. The (f) is the splitting behaviour with
		divergence. See also Figure \ref{spikes} to look the relative spike trains variations.}
	\label{voltage}
\end{figure*}

\section{Results}
The outcome of the simulations show the minimal number of input
sources able to ignite the bump attractor system, and the critical
number of input sources that produce the splitting behaviour of the
spike trains into 2, 3 and 4 streams. An overview of the
spike patterns of the bump attractor are shown in Figure \ref{spikes} (with the relative voltage potential representation in Figure \ref{voltage}).

Table \ref{tab:my-table1} shows the minimal numbers of spike
sources able to ignite the bump attractor network, in relation to the
different weights combinations, that are 1, 2, 4 and 5. The weight combinations related to 1 input are the ones with the highest excitatory weight (0.10) and all the range of inhibitory weights (0.05-0.10). The ignition of the network with 2 inputs is associated with the coupling of many excitatory weights (0.06-0.09) with all inhibitory weights (0.05-0.10). The lowest excitatory weight (0.05) combined with nearly all the inhibitory weights (0.05-0.09) allows the ignition mostly with 4 and 5 inputs. In the case of lowest excitatory weight (0.05) with the highest inhibitory weights (0.10) there is no network ignition even with 40 inputs.

\begin{table}[tbph!]
	\begin{tabular}{|l|l|l|l|l|l|l|}
		\hline
		\textbf{} & \textbf{0.05} & \textbf{0.06} & \textbf{0.07} & \textbf{0.08} & \textbf{0.09} & \textbf{0.1} \\ \hline
		\textbf{0.05} & 4 & 4 & 4 & 4 & 5 & / \\ \hline
		\textbf{0.06} & 2 & 2 & 2 & 2 & 2 & 2 \\ \hline
		\textbf{0.07} & 2 & 2 & 2 & 2 & 2 & 2 \\ \hline
		\textbf{0.08} & 2 & 2 & 2 & 2 & 2 & 2 \\ \hline
		\textbf{0.09} & 2 & 2 & 2 & 2 & 2 & 2 \\ \hline
		\textbf{0.1} & 1 & 1 & 1 & 1 & 1 & 1 \\ \hline
	\end{tabular}
	\caption{Minimal number of spikes sources to ignite the 2-4 bump-attractor network. The $``/"$ means absence of any spikes. }
	\label{tab:my-table1}
\end{table}

\begin{table}[tbph!]
	\begin{tabular}{|c|c|c|c|c|c|c|}
		\hline
		& \textbf{0.05} & \textbf{0.06} & \textbf{0.07} & \textbf{0.08} & \textbf{0.09} & \textbf{0.1} \\ \hline
		\textbf{0.05} & 13 & 13 & 12 & / & / & / \\ \hline
		\textbf{0.06} & 15 & 13 & 13 & 12 & 11 & 11 \\ \hline
		\textbf{0.07} & D & 15 & 14 & 13 & 13 & 12 \\ \hline
		\textbf{0.08} & D & 17(+D) & 15(+D) & 13 & 13 & 13 \\ \hline
		\textbf{0.09} & D & D & D & 15(+D) & 15(+D) & 15 \\ \hline
		\textbf{0.1} & D & D & D & D & D & D \\ \hline
	\end{tabular}
	\caption{Number of inputs that determine the splitting behaviour with 2 streams of the 2-4 bump attractor network. The $ D $ means divergent behaviour and the $ (+D) $ means that that splitting is combined with divergence, whereas $ ``/" $ means absence of any spikes.}
	\label{tab:my-table2}
\end{table}

Table \ref{tab:my-table2} shows
the critical cut-off of inputs that make the splitting behaviour of
the bump net. 
The complete set of combination of excitatory and inhibitory weights ($ 6 \times 6 = 36 $) could be divided in three subsets: 1) the splitting subset, that has splitting behaviour of the
spikes trains (no matter how many streams there are), 2) the divergent
subset, that has divergent behaviour of the spike trains and 3) the
null subset, that does not allow the network to persistently spike.
Approximately, in Table \ref{tab:my-table2}, the splitting subset
is in the upper right triangular part of the matrix,
whereas the divergent subset is in the lower left triangular part of
the matrix. The null subset of weights is within the last three
elements (negative weights equal to 0.08, 0.09 and 0.10) of the first
row (positive weight equal to 0.05).

The splitting subset in Table \ref{tab:my-table2} encompasses the following E/I range:
(0.05/0.05-0.07), (0.06/0.05-0.10), (0.07/0.06-0.10), (0.08/0.06-0.10)
and (0.09/0.08-0.10). It is also sub-divisible in the weights
combinations that allow the emergence of only two streams (Figure
\ref{spikes}-b) and in the combinations that allow the emergence of
three or four streams (see Figure \ref{spikes}-c and Figure
\ref{spikes}-d). The subset with the divergent
behaviour is the following positive and negative combination:
(0.07/0.05), (0.08/0.06), (0.09/0.05-0.07) and (0.10/0.05-0.10). The
patterns having divergent spikes trains have, globally, a common
shape of divergence (see Figure \ref{spikes}-e).

\begin{table}[tbph!]
	\begin{tabular}{|c|c|c|}
		\hline
		\textbf{E-I Weights} & \textbf{3S} & \textbf{4S} \\ \hline
		0.06-0.05 & 25 & 37 \\ \hline
		0.07-0.06 & 26 & 39 \\ \hline
		0.08-0.06 & 26 (+D) & na \\ \hline
		0.08-0.07 & 23 (+D) & na \\ \hline
		0.08-0.08 & 25 & 39 \\ \hline
		0.09-0.08 & 27 (+D) & na \\ \hline
		0.09-0.09 & 25 (+D) & na \\ \hline
	\end{tabular}
	\caption{Combination of excitatory E and inhibitory I weights that determine the splitting behaviour with 3 streams or 4 streams of the 2-4 bump attractor network. The $ (+D) $ means that that splitting is combined with divergence and $ ``na" $ that the pattern is not present in the network configuration.}
	\label{tab:my-table3}
\end{table}

Table \ref{tab:my-table3} shows the combinations of positive and negative weights that
underlie the splitting phenomena with three streams and four
streams, making also a differentiation if splits are merged with the
divergent behaviour of the spike train patterns. For example the weights combination 0.08-0.08 generates the split with 3 streams with 25 (Figure \ref{spikes}-c) inputs and the split with 4 splits with 37 inputs (Figure \ref{spikes}-d), whereas the weights combination 0.09-0.08 generates only a split with 3 streams but also having divergence (see Figure \ref{spikes}-f to observe an example of divergence with streaming).

The next section considers the results achieved and some possible
explanation of the different patterns of spike trains observed in the
bump attractor network.

\section{Discussion}
This paper has explored some critical limits of the number
of inputs sources in a bump attractor network. The criticality is
related to the variability of the spike patterns that the system shows when the input window goes over or under a cut-off.


The principal results from this work regard 1) the minimal number of
spike sources that allow the network to ignite, that is mostly around
2 input sources, excluding the extreme cases with the lowest (0.05)
and highest (0.10) excitatory weights, and 2) the cut-off number of
inputs sources that determines the splitting behaviour with two
streams (see Figure \ref{spikes}-b), that ranges from 11 to 16, with a particular
subset of weights combinations. Avoiding the case when the excitatory
weights are 0.10, the streaming appears when the E/I weights are similar and - as a tendency - when the inhibitory weights are greater than
excitatory weights (that are approximately the upper right triangular part of
the Table \ref{tab:my-table2})).

The stationary spiking pattern (Figure \ref{spikes}-a) emerges when the input
sources window is over the minimal condition for the ignition and
under the critical cut-off for the splitting with two streams (compare
Table \ref{tab:my-table1} with Table \ref{tab:my-table2}).

There is also a collection of weight combinations in the subset of 
the splitting behaviour, that enables the network to have
spike patterns that split with 3 or 4 streams (see Table
\ref{tab:my-table3} and Figure \ref{spikes}-c/d). Note the when three
streams are merged with the divergent behaviour (see Figure
\ref{spikes}-f)). Instead, the whole divergence (see Figure \ref{spikes}-e)) appears in the
complementary subset of weight combinations related to the splitting
behaviour (approximately the lower left triangular part of the Table
\ref{tab:my-table2})

Observing the results obtained, it is possible to make the following
considerations:
\begin{enumerate}
\item  ignition can be achieved
by a few inputs. It is not enough to ignite the network with 
only one spike source, except when the excitatory weight has high
values, as 0.10 (see bottom row of Table \ref{tab:my-table1});

\item the
splitting behaviour with two streams is related to specific
positive and negative weight combinations, that is when they are similar or with greater negative weights than the positive
ones. Therefore, there could be a stronger role for inhibition in the
context of streams genesis, rather than of excitation. To the contrary, the
diverging behaviour of the spike train pattern seems to have a symmetric
explanation, that is the greater role of excitation rather than the
inhibition;

\item the subcases of spike train patterns with 3 and 4 streams and 3
streams with divergence seems related to the size of the input
window. For instance, the number of streams grows as the number of spike
sources increases; therefore the more inputs ignite the bump
network, the more (could be) the streams within the splitting
behaviour (see Table \ref{tab:my-table3});

\item the other subcases of streaming with divergence could be described as
particular weight combination, related to a specific input size, that
share both the property of divergence and splitting behaviour for the
bump network. In particular, the combination of weights that determine
the streaming with divergence are collocated in the boundary between
the weight condition underlying the splitting and
the divergent behaviour of the spike trains. Therefore,
the merged splits and divergence seems an  intermediate
situation of the weights combination close to both the pattern
possibilities.
\end{enumerate}

These results have some limitations. First of all, the topology of the
bump attractor network has  2-4 positive-negative synaptic
connections. Given a neuron, it has excitatory synapses to the
nearest two cells (top and bottom) and inhibitory connections with the
following four cells (top and bottom) (see Figure \ref{bumpNetwork}). Furthermore, it is a one
dimensional network with 100 neurons. A realistic bump attractor
has a relative larger size with a
more flexible positive-negative connection ratio. An example of an anatomical persistent bump attractor is the head direction cells since their activity does
not stop when the light is turned off and the bump is stable in the
absence of input \citep{gerstner2014neuronal}. In this case, the
stimuli or the memory recall operation drives the initial perturbation
of the localized blob of activity, that is a biological bump
attractor. Another example is from the study related to the prefrontal cortex  where the prefrontal persistent activity during the delay of spatial working memory tasks is thought to maintain spatial location in memory. There are recent results in monkey studies (see work by Wimmer et al \cite{wimmer2014bump}) that support a diffusing bump representation for spatial working memory instantiated in persistent prefrontal activity. 

Another limitation is related to the combinatorial exploration of a
predefined set of positive and negative weights coupling, without
using any learning rule, they are basically static weights. In a
realistic bump attractor, there are natural learning strategies, e.g.,
the Hebbian rule \cite{hebb1949}, that govern the weight states in
real time.

Future works will investigate the critical limits of the
bump attractor network taking into account a more realistic
configuration of the topology. Natural learning rules or other weight
combinations could be used. This work adopted a particular
leaky integrate and fire model \citep{gerstner2014neuronal}, but other simulations could regard different neuronal models with other parameter configurations. Regarding the measurement of state
variables, this research focused only on the spike trains and on the voltage membrane potentials without measuring other quantities or deriving interesting ones. 


Further analysis should be done taking account also that the 
bump attractor network is an example of a dynamical system \cite{Meiss:2007,Terman:2008}.
The next computational explorations of the network behaviours should encompass a deeper understanding of those emerging patterns from the theoretical perspective of the physics of complex systems \cite{Nicolis:2007}, applied in the context of the neuronal modelling and simulation by using both standard software environment for brain simulations (e.g. NEST \cite{gewaltig2007nest}) and neuromorphic computing architectures (e.g., SpiNNaker \cite{Furber} and other systems \cite{schuman2017survey}).

\section*{Acknowledgments}
This work was supported by the European Union’s Horizon 2020 research and innovation programme under grant agreement No 720270 (the Human Brain Project).


\bibliographystyle{acm}
\bibliography{nice2020}
\end{document}